\documentclass[letterpaper]{article} 
\usepackage{aaai24}  
\usepackage{times}  
\usepackage{helvet}  
\usepackage{courier}  
\usepackage[hyphens]{url}  
\usepackage{graphicx} 
\usepackage{natbib}  
\usepackage{caption} 
\usepackage{algorithm}
\usepackage{algorithmic}

\usepackage{multirow}

\usepackage{newfloat}
\usepackage{listings}
\usepackage{amsfonts,amsmath,amssymb}

\newcommand{\argmin}[1]{\underset{#1}{\mathrm{argmin}}}

\newtheorem{theo}{Theorem}
\newtheorem{definition}[theo]{Definition}
\newtheorem{prop}[theo]{Proposition}

\newtheorem{lem}[theo]{Lemma}

\newtheorem{assu}[theo]{Assumption}

\usepackage[textwidth=2.5cm, textsize=tiny]{todonotes}

\renewcommand{\arraystretch}{1}
\DeclareCaptionStyle{ruled}{labelfont=normalfont,labelsep=colon,strut=off} 
\floatstyle{ruled}
\newfloat{listing}{tb}{lst}{}
\floatname{listing}{Listing}
\title{A Sequentially Fair Mechanism \\ for Multiple Sensitive Attributes}
\author{
    Fran\c{c}ois Hu\textsuperscript{\rm 1},
    Philipp Ratz\textsuperscript{\rm 2},
    Arthur Charpentier\textsuperscript{\rm 2}
}
\affiliations{
    \textsuperscript{\rm 1}Université de Montréal, Montréal, Québec, Canada\\
    \textsuperscript{\rm 2}Université du Québec à Montréal, Montréal, Québec, Canada\\
    francois.hu@umontreal.ca,
    ratz.philipp@courrier.uqam.ca,  charpentier.arthur@uqam.ca
}

\usepackage{bibentry}

\begin{document}
\maketitle

\begin{abstract}
In the standard use case of Algorithmic Fairness, the goal is to eliminate the relationship between a sensitive variable and a corresponding score. Throughout recent years, the scientific community has developed a host of definitions and tools to solve this task, which work well in many practical applications. However, the applicability and effectivity of these tools and definitions becomes less straightfoward in the case of multiple sensitive attributes. To tackle this issue, we propose a sequential framework, which allows to progressively achieve fairness across a set of sensitive features. We accomplish this by leveraging multi-marginal Wasserstein barycenters, which extends the standard notion of Strong Demographic Parity to the case with multiple sensitive characteristics. This method also provides a closed-form solution for the optimal, sequentially fair predictor, permitting a clear interpretation of inter-sensitive feature correlations. Our approach seamlessly extends to approximate fairness, enveloping a framework accommodating the trade-off between risk and unfairness. This extension permits a targeted prioritization of fairness improvements for a specific attribute within a set of sensitive attributes, allowing for a case specific adaptation. A data-driven estimation procedure for the derived solution is developed, and comprehensive numerical experiments are conducted on both synthetic and real datasets. Our empirical findings decisively underscore the practical efficacy of our post-processing approach in fostering fair decision-making.

\end{abstract}

\section{Introduction}

Recent media coverage has put the spotlight anew on a question that preoccupies the field of (algorithmic) fairness, namely what constitutes fairness and how to achieve it. 
We center our focus on group fairness, particularly on the concept of Demographic Parity fairness~\cite{calders2009building}, with the objective of achieving independence between attributes and predictions while bypassing the use of labels. A point raised frequently is that only considering a single (binary or discrete) attribute is insufficient to determine whether a system is truly fair~\cite{kong2022intersectionally}. Indeed studies emphasize that when a model considers only a single attribute, it overlooks subgroups defined by intersecting attributes. This notion is commonly referred as \textit{fairness gerrymandering}~\cite{kearns2018preventing} and it occurs ``when we only look for
unfairness over a small number of pre-defined groups" that are arbitrarily selected. Scholars in the field of algorithmic fairness have devised a range of methods to counter unfairness in predictions, for both regression~\cite{Chzhen_Denis_Hebiri_Oneto_Pontil20Recali, Chzhen_Denis_Hebiri_Oneto_Pontil20Wasser, gouic2020projection} and classification~\cite{hardt2016equality, agarwal2019fair, chiappa2020general, denis2021fairness}. However, these approaches consider fairness with respect to a single feature, making them susceptible to the criticism from above. As an example, \cite{buolamwini2018gender} underscored this intersectional bias in Machine Learning (ML). They discovered that major face recognition algorithms exhibited preferences for recognizing men and lighter skin tones, resulting in reduced accuracy for women with darker skin tones, thereby exposing both gender and racial discrimination.

A naive solution to this problem would be to create a single discrete feature that groups a set of sensitive variables. However, this approach has several drawbacks: \textit{i)} this methodology assigns similar weights to all attributes, hindering approximate fairness which allows the user to adjust the fairness constraint; \textit{ii)} further, this method complicates tracking procedure effects, as disentangling them from a combined variable is complex and also lacks prioritization across attributes; \textit{(iii)} in applications, some sensitive feature might need more attention, like bias due to gender over age \cite{macnicol2006age,charpentier2023mitigating}. This prioritization bridges the gap between a status-quo of (unfair) predictions and a goal of fair predictions \textit{w.r.t} a set of sensitive features. Fairness considerations often leads to reduced predictive performance \cite{menon2018cost, chen2018my}, making adoption challenging. Presetting a level of unfairness could potentially facilitate its acceptance and adoption.

\subsection{Main Contributions}

This highlights the need for a more holistic approach to consider fairness. In this article, we propose a methodology that is adaptable for optimal fair predictions involving Multiple Sensitive Attributes (MSA). More specifically:
\begin{itemize}
    \item We address the learning problem under the Demographic Parity constraint, involving MSA, by constructing multi-marginal 2-Wasserstein barycenters.
    Our method offers a closed-form solution, allowing us to develop an empirical data-driven approach that enhances fairness for any off-the-shelf estimators.
    \item We rewrite the optimal fair solution into a sequential form by using the associativity of Wasserstein barycenters in univariate settings. This formulation seamlessly extends to approximate fairness achieving fairness-risk trade-off.
    \item Our approach is demonstrated through numerical experiments on diverse datasets (both synthetic and real). It demonstrates high effectiveness in reducing unfairness while enabling clear interpretation of inter-sensitive feature correlations.
\end{itemize}

We begin by introducing some necessary notation before formally presenting the problem. After deriving the main results we conduct extensive numerical experiments and illustrate the use of our methodology on a both synthetic and real-world datasets. Note that all proofs are relegated to the supplementary materials to ease the lecture of the article.

\subsection{Related Work}

Much of our work extends earlier findings from the literature of algorithmic fairness using optimal transport, a mathematical framework for measuring distributional differences. The aim is to transform biased scores into equitable ones while minimizing their impact to uphold predictive performance. There are several methods that can broadly be classified into pre-, in- and post-processing methods. Our approach developed here falls into the latter category, as post-processing is computationally advantageous and allows a clear interpretation of the outputs. In regression, methods like \cite{Chzhen_Denis_Hebiri_Oneto_Pontil20Wasser} and \cite{gouic2020projection} minimize Wasserstein distance to mitigate bias. Similarly, in classification, \cite{chiappa2020general} and \cite{gaucher2023fair} use optimal transport for fairness.
Instead of considering multiple fair attributes, \cite{hu2023fairness} consider multiple fair prediction tasks through joint optimization. However, despite optimal transport's widespread use in algorithmic fairness, there is limited research on MSA and intersectional fairness. This article aims to fill this research gap.

Whereas the majority of recent works in the intersectional fairness domain primarily focus on classification tasks \cite{ijcai2023p742}, this article explores both classification and often-overlooked regression tasks within the intersectional fairness framework. 
Our sequential approach, characterized by the barycenter associativity property, distinguishes itself from conventional methods by enhancing interpretability and naturally addressing inter-correlations among sensitive attributes. Our approach further mitigates data sparsity and uncertainty issues, providing a valuable complementary perspective to existing work in intersectional fairness~\cite{ijcai2023p742}.

Within the MSA framework, multiple studies explore diverse perspectives distinct from ours. \cite{liu2023active} and \cite{jalal2021fairness} use sampling frameworks---one in an active learning setting and the other in image generation---to ensure equal representation of sensitive subgroups. Concurrently, \cite{kilbertus2018blind} addresses disparate impact within privacy constraints without explicit knowledge of sensitive attributes, operating within a fairness-unaware framework. In a similar vein, \cite{hu2023parametric} introduces domain knowledge considerations to alleviate the negative effects of latent sensitive attributes when tackling biases associated with an observed sensitive attribute. In contrast, our method takes a direct approach by adjusting machine learning scores for fairness without the need to select fair samples, distinguishing it from the methodologies employed by \cite{liu2023active} and \cite{jalal2021fairness}. Moreover, our approach operates within a fairness-aware framework, leveraging available sensitive attributes. This sets our method apart from \cite{kilbertus2018blind} and \cite{hu2023parametric}, which operate in fairness-unaware (or semi-unaware) frameworks.

\subsection{Notation}
Consider a function $f$ and a random tuple $(\boldsymbol{X}, \boldsymbol{A})\in\mathcal{X}\times\mathcal{A}\subset\mathbb{R}^d\times \mathbb{N}^r$, with positive integers $d$ and $r$. We denote $\mathcal{V}$ the space of probability measures on $\mathcal{Y}\subset \mathbb{R}$. Let $\nu_{f}\in\mathcal{V}$ and $\nu_{f|\boldsymbol{a}}\in\mathcal{V}$ be respectively the probability measure of $f(\boldsymbol{X}, \boldsymbol{A})$ and $f(\boldsymbol{X}, \boldsymbol{A})|\boldsymbol{A}=\boldsymbol{a}$. $F_{f|\boldsymbol{a}}(u) := \mathbb{P}\left( f(\boldsymbol{X}, \boldsymbol{A}) \leq u|\boldsymbol{A}=\boldsymbol{a} \right)$ corresponds to the cumulative distribution function (CDF) of $\nu_{f|\boldsymbol{a}}$ and $Q_{f|\boldsymbol{a}}(v) := \inf\{u\in\mathbb{R}:F_{f|\boldsymbol{a}}(u)\geq v \}$ its associated quantile function.

\section{Background on Wasserstein Barycenters}

This section introduces the concepts of Wasserstein barycenter from one-dimensional optimal transport theory. Further details can be found in \cite{santambrogio2015optimal, villani2021topics}.

\subsection{Wasserstein Distance} 
We consider two probability measures, $\nu_{1}$ and $\nu_{2}$. The \textit{Wasserstein distance} quantifies the minimum "cost" of transforming one distribution into the other. Specifically, the squared Wasserstein distance between $\nu_{1}$ and $\nu_{2}$ is defined as
\begin{equation*}
    \mathcal{W}_2^2(\nu_{1}, \nu_{2}) = \inf_{\pi\in\Pi(\nu_{1}, \nu_{2})} \mathbb{E}_{(Z_1, Z_2)\sim \pi}\left(Z_2-Z_1\right)^2 \enspace,
\end{equation*}
where $\Pi(\nu_{1}, \nu_{2})$ is the set of distributions on $\mathcal{Y}\times\mathcal{Y}$ having $\nu_{1}$ and $\nu_{2}$ as marginals. A coupling that achieves this infimum is called optimal coupling between $\nu_{1}$ and $\nu_{2}$. 

\subsection{Wasserstein Barycenter}
Throughout this article, we will frequently make use of \textit{Wasserstein Barycenters} \cite{agueh2011barycenters}. The Wasserstein barycenter finds a representative distribution that lies between multiple given distributions in the Wasserstein space. It is defined for a family of $K$ measures $(\nu_1, \dots, \nu_K)$ in $\mathcal{V}$ and some positive weights $(w_1, \dots, w_K) \in\mathbb{R}_+^K$. The Wasserstein barycenter (or in short $\mathcal{W}_2$-barycenter), denoted as ${\rm Bar}\left\{(w_k, \nu_k)_{k=1}^K\right\}$ is the minimizer
\begin{equation}\label{eq:WB}
    {\rm Bar}(w_k, \nu_k)_{k=1}^K = \argmin{\nu}\sum_{k=1}^{K}w_k\cdot \mathcal{W}_2^2\left( \nu_k, \nu\right)\enspace.
\end{equation}
The work in \cite{agueh2011barycenters} shows that in our configuration the barycenter exists and a sufficient condition of uniqueness is that one of the measures $\nu_i$ admits a density w.r.t. the Lebesgue measure.

Our study focuses on barycentric associativity within the unidimensional space $\mathcal{Y}\subset\mathbb{R}$. This principle asserts that the global barycenter coincides with the barycenter of barycenters. This associativity is clear in Euclidean spaces~\cite{ungar2010barycentric}: the barycenter of points $x_1$, $x_2$, and $x_3$ in $\mathbb{R}^d$ with weights $w_1$, $w_2$, and $w_3$ aligns with the barycenter of $x_{1, 2}$ and $x_3$ with weights $w_1+w_2$ and $w_3$, where $x_{1, 2}$ is the barycenter of $x_1$ and $x_2$. In the following proposition, we explore the relevance of this barycentric associativity, particularly for the 2-Wasserstein barycenter within the unidimensional space. It is important to note that the Wasserstein barycenter loses its associativity in a multi-dimensional framework.

\begin{prop}[Associativity of the $\mathcal{W}_2$-barycenter]\label{prop:MainPropBary}
    Consider a collection of positive integers $K_1, K_2, \dots, K_r$, where their sum is denoted as $K = K_1 + K_2 + \dots + K_r$. Let the sets be defined as follows:
    $$
    B_1 = (w_{1, k}, \nu_{1, k})_{k=1}^{K_1}, \ldots, B_r = (w_{r, k}, \nu_{r, k})_{k=1}^{K_r}\enspace,$$
    where $\{w_{i, k}\}_{i, k}$ are positive and non-zero weights summing to 1 and $\{\nu_{i,k}\}_{i,k}$ represent univariate measures. In this context, the overall Wasserstein barycenter ${\rm Bar}\left\{B_1 \cup \cdots \cup B_r\right\}$ can be expressed as the barycenter $
    {\rm Bar} (\Tilde{w}_{i}, {\rm Bar}(\Tilde{B}_i))_{i=1, \ldots, r}$, where
$$
\Tilde{w}_{i} := \sum_{k'=1}^{K_i}w_{i,k'} \ 
\text{ and } \
\Tilde{B}_i := \left(\frac{w_{i,k}}{\Tilde{w}_i}\ ,\  \nu_{i,k}\right)_{k=1, \ldots, K_i}\enspace.
$$
\end{prop}
In a nutshell, this formulation captures the relationship between the overall Wasserstein barycenter and the individual barycenters of its constituent sets, incorporating the relevant weights seamlessly. 
Given measures $(\nu_1, \nu_2, \nu_3)$ with positive weights $(w_1, w_2, w_3)$ summing to 1, mirroring the aforementioned barycentric concept in Euclidean space, we derive the following relations:
\begin{multline*}
{\rm Bar}\left\{(w_1, \nu_1), (w_2, \nu_2), (w_3, \nu_3)\right\} \\=  {\rm Bar}\left\{(w_1,\nu_1), (w_2+w_3,\nu_{2,3})\right\}
\\=  {\rm Bar}\left\{(w_1+w_2,\nu_{1,2}), (w_3,\nu_3) \right\}\enspace.
\end{multline*}
Here, for $i, j\in \{1, 2, 3\}$ (where $i\neq j$), the measure $\nu_{i,j} = {\rm Bar}\left\{(\Tilde{w}_i, \nu_i), (\Tilde{w}_j, \nu_j)\right\}$ is defined, with $\Tilde{w}_k = w_k/(w_i+w_j)$.



\section{Problem Formulation}

Let $(\boldsymbol{X}, \boldsymbol{A}, Y)$ be a random tuple with distribution $\mathbb{P}$. Here, $\boldsymbol{X}\in\mathcal{X}\subset \mathbb{R}^d$ denotes the $d$ non-sensitive features, $Y\in \mathcal{Y} \subset \mathbb{R}$ represents the target task, and $\boldsymbol{A}=(A_1,\ldots, A_r) \in \mathcal{A} := \mathcal{A}_1\times \cdots\times\mathcal{A}_r$ the $r$ discrete sensitive attributes, where $\mathcal{A}_i = \{1, \ldots, K_i\}$ with $K_i\in\mathbb{N}$. For example, in a binary case with $r=2$, we could have $A_1 = \text{gender}$ and $A_2 = \text{age}$.
For convenience, we use the notation $A_{i:i+k} := (A_i, A_{i+1}, \cdots, A_{i+k})$ to denote the sequence of $k+1$ sensitive features ranging from $i$ to $i+k$ (so $\boldsymbol{A} = A_{1:r}$). Further, we denote $\mathcal{F}$ the set of predictors of the form $f:\mathcal{X}\times\mathcal{A}\to \mathcal{Y}$ where we assume that each measure $\nu_{f|\boldsymbol{a}}$ admits a density w.r.t. Lebesgue measure. More precisely, we require the following assumption to hold:
\begin{assu}\label{assu:general}
    Given $f\in\mathcal{F}$, measures $\{\nu_{f|\boldsymbol{a}}\}_{\boldsymbol{a}\in\mathcal{A}}$ are non-atomic with finite second moments.
\end{assu}

\subsection{Risk Measure}

Within the statistical learning community, a central pursuit revolves around the minimization of a designated risk measure across the set $\mathcal{F}$ encompassing all predictors. In particular, a Bayes regressor minimizing the squared risk,
$$
\text{\textbf{(Risk Measure)}}\quad \mathcal{R}(f) := \mathbb{E} \left( Y - f(\boldsymbol{X}, \boldsymbol{A}) \right)^2\enspace,
$$
over the set $\mathcal{F}$ is represented by $f^*(\boldsymbol{X}, \boldsymbol{A}) := \mathbb{E}[Y|\boldsymbol{X}, \boldsymbol{A}]$.

In our case, our objective is to characterize the optimal fair predictor, which minimizes the squared risk under a given fairness constraint. To do so, we introduce formally the Demographic Parity notion of fairness.


\subsection{Unfairness Measure}
Demographic Parity (DP) will be used to determine the fairness of a predictor. 
Fairness considerations under DP offers the advantage of being applicable to both classification and regression tasks.
In our study, the unfairness measure of the predictor on the feature $A_i$ is given by

\begin{equation}\label{eq:UnfairnessSingle}
\mathcal{U}_i(f) = \max_{a_i\in \mathcal{A}}\int_{u\in[0,1]} \left|\ Q_{f}(u) - Q_{f|a_i}(u)\ \right|du\enspace,
\end{equation}
while for the multiple sensitive features $A_{i}, \dots,A_{i+k}$, their collective unfairness is simply assessed through:
\begin{equation}\label{eq:UnfairnessMulti}
    \mathcal{U}_{\{i, \dots, i+k\}}(f) = \mathcal{U}_{i:i+k}(f) = \mathcal{U}_i(f) + \dots + \mathcal{U}_{i+k}(f)\enspace.
\end{equation}
Hence, we naturally broaden the DP-fairness definition to encompass both exact and approximate fairness within the context of MSA framework.

\begin{definition}[Fairness under Demographic Parity]\label{def:DPFair}
The overall unfairness of a predictor $f\in\mathcal{F}$ \textit{w.r.t.} the feature $\boldsymbol{A} = A_{1:r}$, can be quantified by the unfairness measure,
$$
\text{\textbf{(Unfairness measure)}} \quad \mathcal{U}(f) = \mathcal{U}_{1:r}(f)\enspace.
$$
Then $f$ is called exactly fair if and only if 
$$
\mathcal{U}(f) = 0\enspace.
$$
Given $\boldsymbol{\varepsilon} = \varepsilon_{1:r} := (\varepsilon_1, \varepsilon_2, \ldots, \varepsilon_r)$ where each $\varepsilon_i \in [0, 1]$, $f$ is called approximately fair under DP with $\boldsymbol{\varepsilon}$ relative improvement ($\boldsymbol{\varepsilon}$-RI) if and only if each individual unfairness satisfies
$$
\mathcal{U}_i(f) \leq  \varepsilon_i\times \mathcal{U}_i(f^*)\enspace.
$$
\end{definition}
In other words, in the context of approximate fairness, we are interested in the relative (fairness) improvement of a fair predictor with respect to Bayes' rule $f^*$ (see \cite{chzhen2022minimax} for further details).

\subsection{Preliminary Results}

Recall the Wasserstein barycenter defined in Eq.~\eqref{eq:WB}. We consider measures $(\nu_{f|\boldsymbol{a}})_{\boldsymbol{a}\in\mathcal{A}}$ with corresponding weights $(p_{\boldsymbol{a}})_{\boldsymbol{a}\in\mathcal{A}}$, where $p_{\boldsymbol{a}} := \mathbb{P}(\boldsymbol{A}=\boldsymbol{a})$. It is assumed that $\min_{\boldsymbol{a}}\{p_{\boldsymbol{a}}\} \geq 0$. Encompassing these measures is the Wasserstein barycenter, denoted $\mu_{\mathcal{A}}:\mathcal{V}\to \mathcal{V}$ and it is defined as
\begin{multline*}
    \mu_{\mathcal{A}}(\nu_f) := {\rm Bar}(p_{\boldsymbol{a}}, \nu_{f|{\boldsymbol{a}}})_{{\boldsymbol{a}}\in\mathcal{A}} \\ = \argmin{\nu}\sum_{\boldsymbol{a}\in\mathcal{A}}p_{\boldsymbol{a}}\cdot \mathcal{W}_2^2\left( \nu_{f|{\boldsymbol{a}}}, \nu\right)\enspace.
\end{multline*}


\subsubsection{Single Sensitive Attribute (SSA) Case} 
We consider a single sensitive attribute $A$, belonging to the set $\mathcal{A} = \{1, \dots, K\}$, with $p_a := \mathbb{P}(A=a)$. Let $f_B\in\mathcal{F}$, and let its measure be the Wasserstein barycenter $\nu_{f_B} = \mu_{\mathcal{A}}(\nu_{f^*})$, where we recall that $f^*(\boldsymbol{X}, A) = \mathbb{E}[Y|\boldsymbol{X}, A]$ is the Bayes rule which minimizes the squared risk. Prior research conducted by \cite{Chzhen_Denis_Hebiri_Oneto_Pontil20Wasser,gouic2020projection} demonstrates that,
\begin{equation*}
    f_B = \argmin{f\in\mathcal{F}} \left\{\mathcal{R}(f) : \mathcal{U}(f) = 0\right\}\enspace.
\end{equation*}
Therefore, $f_B$ represents the optimal fair predictor in terms of minimizing unfairness-risk. Additionally, previous studies have offered a closed-form solution: for all $(\boldsymbol{x}, a) \in \mathcal{X}\times \mathcal{A}$,
\begin{equation*}
    f_B(\boldsymbol{x}, a) = \left( \sum_{a'\in\mathcal{A}} p_{a'}Q_{f^*|a'}\right)\circ F_{f^*|a}\left( f^*(\boldsymbol{x}, a) \right)\enspace.
\end{equation*}
Note that this solution can be easily adapted not only to classification tasks \cite{gaucher2023fair}, but also to multi-task learning involving classification and regression \cite{hu2023fairness}. In this article, we extend this formulation to the MSA case $\boldsymbol{A} = (A_1, \dots, A_r)\in\mathcal{A} = \mathcal{A}_1\times\dots \times \mathcal{A}_r$. For any $(\boldsymbol{x}, \boldsymbol{a})\in \mathcal{X}\times \mathcal{A}$ we denote,
\begin{equation}\label{eq:OptFairi}
    f_{B_i}(\boldsymbol{x}, \boldsymbol{a}) = \left( \sum_{a_i'\in\mathcal{A}_i} p_{a_i'}Q_{f^*|a_i'}\right)\circ F_{f^*|a_i}\left( f^*(\boldsymbol{x}, \boldsymbol{a}) \right)\enspace,
\end{equation}
as the optimal fair predictor, ensuring fairness only across $\mathcal{A}_i$ ($\mathcal{A}_i$-fair for short). By abuse of notation,
we denote $p_{a_i'} := \mathbb{P}(A_i=a_i')$, $F_{f^*|a_i'}(u):= \mathbb{P}\left( f(\boldsymbol{X}, \boldsymbol{A}) \leq u|A_i=a_i'\right)$ and $Q_{f^*|a_i'}$ its associated quantile function.



\section{Optimal Fair Prediction with MSA}
We extend the fair characterization into a sequential framework to accommodate MSA. Building upon previous research in the SSA case \cite{Chzhen_Denis_Hebiri_Oneto_Pontil20Wasser, gouic2020projection}, we demonstrate that fairness in the MSA problem can also be framed as the optimal transport problem involving the 2-Wasserstein distance. The relationship between these concepts is established in the following proposition.

\begin{prop}[Fair characterization: global approach]
\label{prop:MainFairOptimal}
We assume that Assumption~\ref{assu:general} holds, and we let
\begin{equation*}
    f_B = \argmin{f\in\mathcal{F}} \left\{\mathcal{R}(f) : \mathcal{U}(f) = 0\right\}\enspace.
\end{equation*}
Subsequently, its measure satisfies $\nu_{f_B} = \mu_{\mathcal{A}}(\nu_{f^*})$.
Furthermore, this equation yields a closed-form solution for the optimal fair predictor
\begin{equation}\label{eq:OptFairGlobal}
    f_B(\boldsymbol{x}, \boldsymbol{a}) = \left( \sum_{\boldsymbol{a}'\in\mathcal{A}} p_{\boldsymbol{a'}}Q_{f^*|\boldsymbol{a}'}\right)\circ F_{f^*|\boldsymbol{a}}\left( f^*(\boldsymbol{x}, \boldsymbol{a}) \right)\enspace.
\end{equation}
\end{prop}

Considering the aforementioned proposition and Prop.~\ref{prop:MainPropBary}, which confirms that the barycenter of barycenters aligns with the overall barycenter under suitable updated weights, a straightforward corollary arises. This corollary asserts that regardless of the selected ``debiasing path" in the sequential fairness mechanism, the end result consistently leads to the same optimal fair solution.

\begin{prop}[Sequentially fair mechanism]\label{prop:SFairOptimal}
Under the assumption that Assumption~\ref{assu:general} holds, the term $\mu_{\mathcal{A}}(\nu_{f^*})$ defined in Prop.~\ref{prop:MainFairOptimal} can be equivalently expressed as follows:
\begin{equation*}
    \mu_{\mathcal{A}}(\nu_{f^*}) = \mu_{\mathcal{A}_1}\circ\mu_{\mathcal{A}_2}\circ\cdots\circ \mu_{\mathcal{A}_r}(\nu_{f^*})\enspace.
\end{equation*}
More generally, under any permutation, i.e. bijection function of the form $\sigma : \mathcal{S} \to \mathcal{S}$ where $\mathcal{S} := \{1, 2, \ldots, r\}$, the above expression can be rewritten as
\begin{equation*}
    \mu_{\mathcal{A}}(\nu_{f^*}) = \mu_{\mathcal{A}_{\sigma(1)}}\circ\mu_{\mathcal{A}_{\sigma(2)}}\circ\cdots\circ \mu_{\mathcal{A}_{\sigma(r)}}(\nu_{f^*})
    \enspace.
\end{equation*}

\end{prop}
Notably the expressions proposed in Prop.~\ref{prop:SFairOptimal} allows us to establish a link between Eq.~\eqref{eq:OptFairi} and Eq.~\eqref{eq:OptFairGlobal} of the form:
\begin{align*}
f_{B} (\boldsymbol{X}, \boldsymbol{A}) &= \left(f_{B_1}\circ f_{B_2}\circ \ldots \circ f_{B_r}\right)(\boldsymbol{X},\boldsymbol{A})\\
&= \left(f_{B_{\sigma(1)}}\circ f_{B_{\sigma(2)}}\circ \ldots \circ f_{B_{\sigma(r)}}\right)(\boldsymbol{X}, \boldsymbol{A})\enspace.
\end{align*}
Note that the $\circ$ notation is used in a relaxed manner with regard to predictors, aimed at streamlining the presentation and alleviating the complexities of notation. In particular, by abuse of notation we establish the definition of $f_{B_i}\circ f_{B_j}$ as follows (with $f^*$ serving as the default function):
    \begin{multline*}
        \left( f_{B_i}\circ f_{B_j} \right)(\boldsymbol{x}, \boldsymbol{a}) \\ = \left( \sum_{a_i'\in\mathcal{A}_i} p_{a_i'}Q_{f_{B_j}|a_i'}\right)\circ F_{f_{B_j}|a_i}\left( f_{B_j}(\boldsymbol{x}, \boldsymbol{a}) \right)\enspace.
    \end{multline*}

Introducing a sequential approach is pivotal for enhancing clarity. Indeed, this methodology helps in comprehending intricate concepts like approximate fairness with $\boldsymbol{\varepsilon}$-RI defined in Definition~\ref{def:DPFair}, which involves improving fairness relatively and approximately with $\boldsymbol{\varepsilon}$ a preset level of relative fairness improvement. This perspective enables us to break down how various components of the sequential fairness mechanism interact to achieve fairness goals and allows for the interpretation of the intrinsic effects of adjusting fairness. Therefore, adopting a sequential perspective is a key step in attaining a deeper understanding of fairness and bias.



\section{Extension to Approximate Fairness}

Recall the previously introduced concept of $\boldsymbol{\varepsilon}$-RI fairness (or $\boldsymbol{\varepsilon}$-fairness for brevity), where $\boldsymbol{\varepsilon} = \varepsilon_{1:r} = (\varepsilon_{1}, \cdots, \varepsilon_{r})$ and each $\varepsilon\in[0, 1]$. In the context of the SSA framework, a methodology introduced by \cite{chzhen2022minimax} employs geodesic parameterization. Specifically, considering $A_1\in\mathcal{A}_1$ \textit{w.l.o.g.}, the predictor of the form:
$$
f_{B_1}^{\varepsilon_1}(\boldsymbol{X}, \boldsymbol{A}) = (1-\varepsilon_1)\cdot f_{B_1}(\boldsymbol{X}, \boldsymbol{A}) + \varepsilon_1 \cdot f^*(\boldsymbol{X}, \boldsymbol{A})\enspace,
$$
achieves the optimal risk-fairness trade-off. Notably, we have
$$
f_{B_1}^{\varepsilon_1} \in \argmin{f\in\mathcal{F}}\{ \mathcal{R}(f): \mathcal{U}_1(f)\leq\varepsilon_1 \cdot \mathcal{U}_1(f^*)\}\enspace.
$$
We denote the corresponding measure as $\mu_{\mathcal{A}_1}^{\varepsilon_1}(\nu_{f^*}) := \nu_{f_{B_1}^{\varepsilon_1}}$. In the following proposition, we extend sequentially this formulation to the context of MSA.

\begin{prop}[Characterization of approximate fairness]\label{prop:SFairOptimalApprox}
    Let Assumption~\ref{assu:general} holds and let
    $$
f_{B}^{\boldsymbol{\varepsilon}} = \argmin{f\in\mathcal{F}}\left\{ \mathcal{R}(f): \mathcal{U}(f) \leq \sum_{i=1, \ldots, r}\varepsilon_i \cdot \mathcal{U}_i(f^*) \right\}\enspace,
    $$
then 
$$
\nu_{f_{B}^{\boldsymbol{\varepsilon}}} = \mu_{\mathcal{A}}^{\boldsymbol{\varepsilon}}(\nu_{f^*}) := \mu_{\mathcal{A}_1}^{\varepsilon_1}\circ\cdots\circ \mu_{\mathcal{A}_r}^{\varepsilon_r}(\nu_{f^*})
\enspace.
$$
Similarly to Prop.~\ref{prop:SFairOptimal}, this expression can also be reformulated by permuting indices.
\end{prop}
The expression mentioned in Prop.~\ref{prop:SFairOptimalApprox} enables us to explicitly formulate an optimal closed-form predictor within the approximate fairness framework: for any permutation $\sigma\in\mathcal{P}(\mathcal{S})$,
\begin{equation*}
f^{\boldsymbol{\varepsilon}}_{B} (\boldsymbol{X}, \boldsymbol{A}) =  \left(f^{\varepsilon_{\sigma(1)}}_{B_{\sigma(1)}}\circ f^{\varepsilon_{\sigma(2)}}_{B_{\sigma(2)}}\circ \ldots \circ f^{\varepsilon_{\sigma(r)}}_{B_{\sigma(r)}}\right)(\boldsymbol{X}, \boldsymbol{A})\enspace.
\end{equation*}

\begin{figure*}[!h]
\centerline{\includegraphics[width=0.82\textwidth]{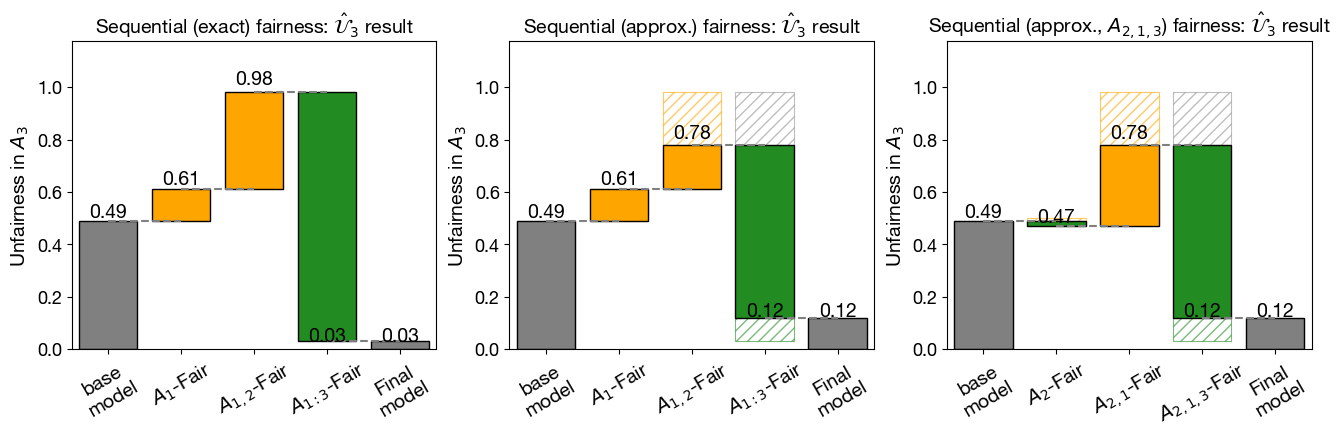}}
\caption{Synthetic data with $\boldsymbol{\tau} = (0, 0.05, 0.1)$. A sequential unfairness evaluation, $\mathcal{U}_3$, of (\textit{left pane}) exact fairness, (\textit{middle}) approximate $A_{1:3}$-fairness with $\boldsymbol{\varepsilon}$-RI where $\boldsymbol{\varepsilon} = \varepsilon_{1, 2, 3} = (0.2, 0.5, 0.75)$ and (\textit{right}) approximate $A_{2, 1, 3}$-fairness with $\varepsilon_{2, 1, 3}$-RI.
}
\label{fig:waterfall_syn_unfairnessA3}
\end{figure*}

\begin{figure*}[h!]
\centerline{\includegraphics[width=0.71\textwidth]{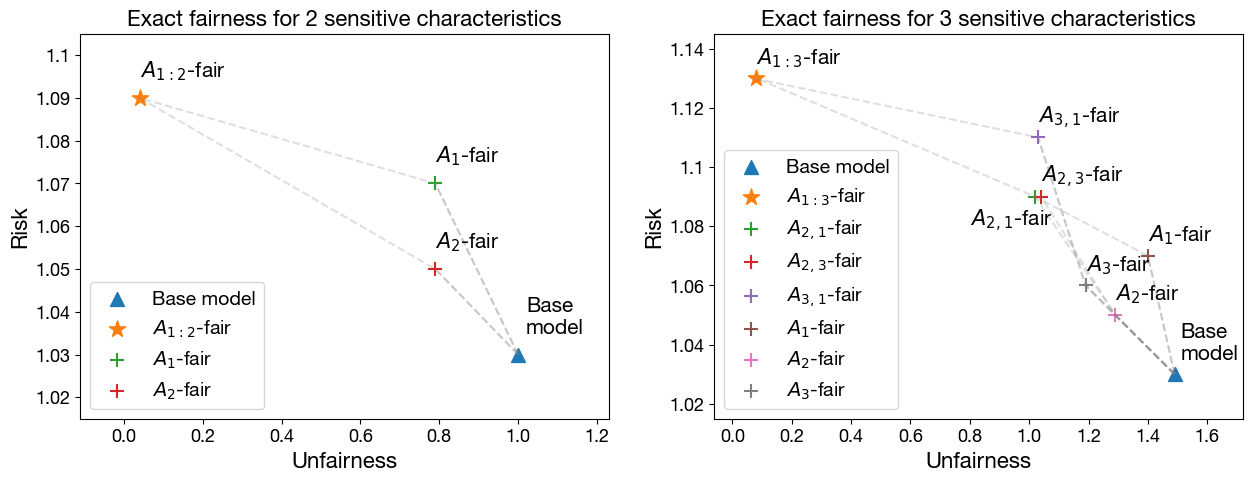}}
\caption{(Risk, Unfairness) phase diagrams that shows the sequential fairness approach for (\textit{left}) two and (\textit{right}) three sensitive features. In this study, Unfairness represents the overall unfairness $\Hat{\mathcal{U}} =\Hat{\mathcal{U}}_{1:3}$. Bottom-left corner gives the best trade-off.}
\label{fig:line_syn_tradeoff}
\end{figure*}

\section{Data-Driven Procedure}


For practical application on real data, the plug-in estimator of the Bayes rule $f^*$ is denoted as $\hat{f}$—any DP-unconstrained ML model trained on a set \textit{i.i.d.} instances of $(\boldsymbol{X}, \boldsymbol{A}, Y)$. Given $\boldsymbol{x}\in\mathcal{X}$ and $\boldsymbol{a}=(a_1, \ldots, a_r)\in\mathcal{A}$, the empirical counterpart of an optimal $\mathcal{A}_i$-fair predictor $f_{B_i}$ is then defined as:
\begin{equation}\label{eq:plugin}
  \widehat{f_{B_i}}(\boldsymbol{x}, \boldsymbol{a}) = \left( \sum_{a_i'\in\mathcal{A}_i} \Hat{p}_{a_i'}\Hat{Q}_{\Hat{f}|a_i'}\right)\circ \Hat{F}_{\Hat{f}|a_i}\left( \Hat{f}(\boldsymbol{x}, \boldsymbol{a}) \right)\enspace,  
\end{equation}
Here, $\Hat{p}_{a_i}$, $\Hat{F}_{\hat{f}|a_i}$, and $\Hat{Q}_{\hat{f}|a_i}$ are empirical counterparts of $p_{a_i}$, $F_{f^*|a_i}$, and $Q_{f^*|a_i}$. Interestingly, aside from $\hat{f}$, the other quantities can be constructed using an unlabeled dataset. Notably, \cite{Chzhen_Denis_Hebiri_Oneto_Pontil20Wasser} provides some statistical guarantees: if the estimator $\Hat{f}$ approximates $f^*$ well, then given mild assumptions on distribution $\mathbb{P}$, the post-processing method $\widehat{f_{B_i}}$ is a good estimator of $f_{B_i}$. By composition, $\widehat{f_{B}} = \widehat{f_{B_1}}\circ ... \circ \widehat{f_{B_r}}$ and $\widehat{f^{\boldsymbol{\varepsilon}}_{B}} = \widehat{f^{\varepsilon_1}_{B_1}}\circ ... \circ \widehat{f^{\varepsilon_r}_{B_r}}$ emerge as good estimators for  $f_{B}$ and $f^{\boldsymbol{\varepsilon}}_{B}$ respectively, enabling accurate and fair estimation of the instances. Note that the unfairness of $f$ is assessed on a test set using $\widehat{\mathcal{U}}(f)$, the empirical counterpart of Eq.~\eqref{eq:UnfairnessMulti}. Predictive performance uses mean squared error (MSE) for regression and $F_1$-score for classification on the same test set.

\section{Numerical Experiments}

In this section, we conduct a comparative analysis of our methodology against DP-unconstrained methods and state-of-the-art approach given in \cite{agarwal2019fair}. Our findings demonstrate that our exact and approximate fairness approach stands out in terms of interpretability, adaptability and competitiveness.

\subsection{Case Study on Synthetic Data}


Prior to showcasing our method on a real dataset, we opted to assess its performance using synthetic data. This step aims to provide a clearer insight into the effectiveness of the sequential fairness mechanism. Specifically, we consider synthetic data $(\boldsymbol{X}, \boldsymbol{A}, Y)$ with the following characteristics:

\begin{itemize}
    \item $\boldsymbol{X}\in\mathbb{R}^d$: Comprises $d$ non-sensitive features generated from a centered Gaussian distribution $\boldsymbol{X}\sim \mathcal{N}_{d}(0, \sigma_X I_d)$, where $\sigma_X > 0$ parameterizes its variance.
    \item $\boldsymbol{A}=A_{1:r} \in\{-1, 1\}^r$: Represents $r$ binary sensitive features, with $A_i\sim 2\cdot\mathcal{B}(q_i)-1$ following a Bernoulli law with parameter $q_i = \mathbb{P}(\Tilde{X} > \tau_i)$, where $\Tilde{X}\sim \mathcal{N}(0, \sigma_X)$. Here, $\boldsymbol{\tau} = (\tau_1, \ldots, \tau_r)$ is a user-set parameter.
    \item $Y \sim \mathcal{N}(\boldsymbol{1}^T\boldsymbol{X} + \boldsymbol{1}^T\boldsymbol{A}, 1) $: Represents the regression task.
\end{itemize}
\subsubsection{Simulation Scheme}
Default parameters are set as follows: $d = 10$, $r=3$, $\sigma_X = 0.15$, and $\boldsymbol{\tau} = (0, 0.05, 0.1)$. We generated 10,000 synthetic examples and divided the data into three sets (50\% training, 25\% testing, and 25\% unlabeled). As a base model, we opt for a simple linear regression using default parameters from \texttt{scikit-learn} in Python. Comparatively, we assess our sequential approach against the uncalibrated base model.

\subsubsection{Interpreting Intersectional Fairness}
In the context of $r=3$ sensitive features, Figure~\ref{fig:waterfall_syn_unfairnessA3} showcases $\Hat{\mathcal{U}}_3$, an unfairness measure focusing on $A_3$. The sequential fair approach, detailed in Prop. \ref{prop:SFairOptimal}, enhances interpretability by addressing inter-correlations among sensitive features while striving for fair predictions. Specifically, this highlights the $A_1$ and $A_2$ correlation with $A_3$. Pursuing exact fairness, the left side of Figure \ref{fig:waterfall_syn_unfairnessA3} demonstrates that making $A_1$ and $A_2$ fair can inadvertently introduce unfairness in $A_3$, revealing the fairness gerrymandering issue mentioned in the introduction. Aligning with Prop.~\ref{prop:SFairOptimal} and Prop.~\ref{prop:SFairOptimalApprox}, both Figure~\ref{fig:waterfall_syn_unfairnessA3} and Figure~\ref{fig:line_syn_tradeoff} exhibit varied numerical debiasing paths to achieve fairness across the three sensitive features. Importantly, this implies that achieving fairness for $A_1$ before $A_2$ is equivalent to the reverse, evident in Figure~\ref{fig:line_syn_tradeoff}'s (Risk, Unfairness) phase diagrams. 
Each step illustrated incurs a performance loss but garners fairness gains. The number of such points is influenced by the cardinality of the power set encompassing all sensitive features. Supplementary details on additional fairness experiments with this synthetic data are available in the supplementary materials.

\subsection{Case Study with Two Sensitive Attributes}

\begin{figure*}[htbp]
\centerline{\includegraphics[width=0.84\textwidth]{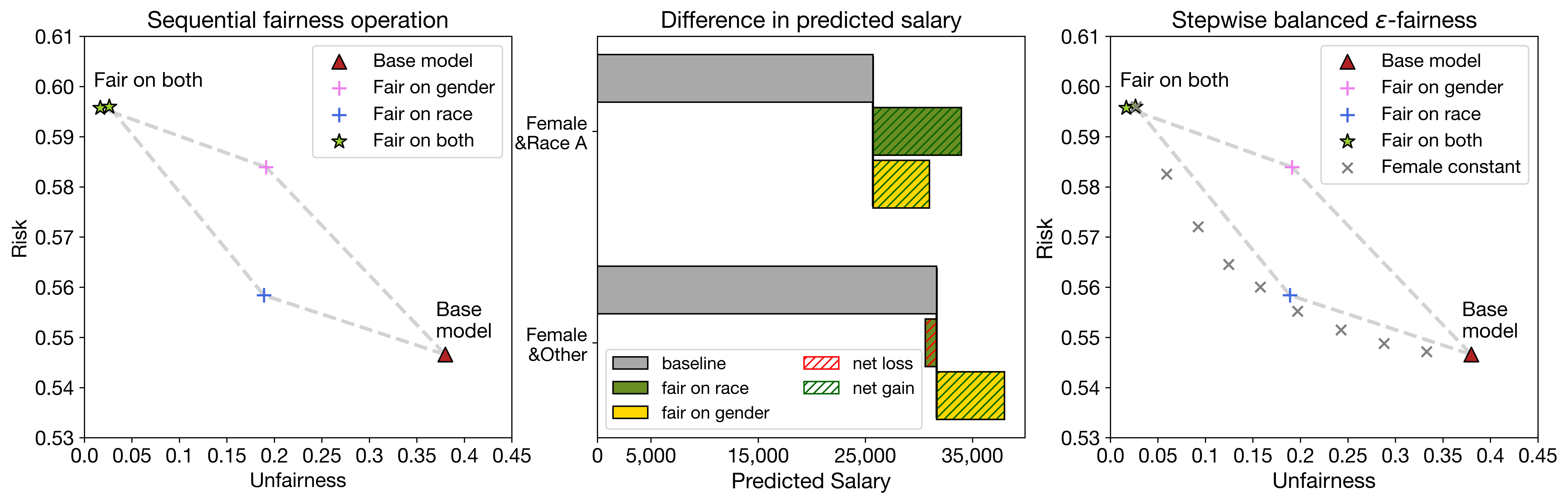}}
\caption{Applications on the \emph{folktables} data set. Left pane, visualisation of the combined unfairness across two sensitive attributes and intermediate solutions rendering predictions fair on only one of them. Center pane, marginal changes to predicted income when rendering fair the predictions w.r.t. a single variable and the baseline predictions. Right pane, visualization of global metrics when correcting the score first for race, but keeping the average predicted salary of female individuals constant.}
\label{fig:application_all_figs}
\end{figure*}

To illustrate a possible application of our methodology and showcase its use in a real world use-case, we consider data collected in the \emph{folktables} package of \cite{ding2021retiring}. This package compiles datasets sourced from the US Census, offering a basis for benchmarking ML models. Notably, with named features, we can add interpretation to our methodology. Our study centers on predicting an individual's total income (Income) within sunbelt states (AL, AZ, CA, FL, GA, LA, MS, NM, SC, TX) using standard filters (age $\geq 18$, at least one hour of weekly work, total income $>100$\$). As a secondary task, we consider the problem of predicting whether an individual is covered by public health insurance (Coverage), again with the standard filters (age $<$ 65, income $\leq$ 30,000\$). For both prediction problems, we rely on provided standard features and treat \textit{gender} and \textit{race} as sensitive attributes\footnote{Code available at: \url{https://github.com/phi-ra/SequentialFairness}.}.



\subsubsection{Methodology}
As our approach is applicable to both regression and classification tasks, we consider both problems. For the regression task, we aim to predict the log-income and for the classification task, we aim to predict whether an individual's income exceeds 50,000\$. The sensitive features studied are \emph{gender} and \emph{race}. In total, we have 600,041 observations, with 52.3\% Male participants and 10.7\% participants from the minority racial class. The data is split into 64\% train, 20\% test and 16\% unlabeled  data. As a base model, we opt for a lightGBM \cite{NIPS2017_6449f44a} model with early stopping, where the early stopping iterations are optimized using 5-fold cross validation on the training data.

In an exact fairness framework ($\boldsymbol{\varepsilon}=0$), we evaluate numerical performance over ten runs with distinct seeds. Since, to our knowledge, there are no open-source fairness methods for MSA predictions, direct benchmarking is not easily achieved. For the main classification task, we compare our approach with the state-of-the-art \emph{ExponentiatedGradient} method from the \emph{Fairlearn} package \cite{weerts2023fairlearn}. First, we ensure fairness on a single variable, enabling a comparison of baseline $F_1$-Score, unfairness and compute-time. Next, our methodology is employed to ensure fairness across both variables, facilitating another comparison. To create further performance comparisons, we also run benchmarks on well-known data sets with differing number of sensitive variables ($s$ in Table \ref{tab1}) using \emph{Fairlearn} and \emph{FairBalance} from \cite{yu2021fairer}. Here, we create sensitive hyper-features to allow easier learning for the two benchmarking methods. This should help them learn a fair representation but will likely increase compute time.

\subsubsection{Results}

From Table~\ref{tab1}, in regression, the outcomes align with expectations. Post-processed forecasts exhibit slightly reduced predictive performance, yet offer nearly complete fairness. This achievement comes with minimal added computational cost, quantified by seconds of compute time. In classification, our method proves competitive against the benchmark model. Indeed, both methods provide efficient and fair outcomes for a single feature, though the post-processing of our method is significantly faster. With two sensitive features, the benchmark model underperforms in fairness (as anticipated), while our method excels across both tasks. Though performance slightly dips compared to a single sensitive feature, competitiveness persists—even against the benchmark model, which fails to achieve fairness across both sensitive features. The last three sections in Table \ref{tab1} contain the further performance benchmarks. Across all experiments, our method performs consistently fastest and best on the Fairness metric and competitive on the performance metric.

\begin{table}[htbp]
\footnotesize
\begin{center}
\def\arraystretch{1.25}%
\begin{tabular}{|c c c c|}
\hline
  & \textbf{\textit{Uncorrected}} & \textbf{\textit{Our Method}} & \textbf{\textit{Fairlearn}} \\
\hline
 & \multicolumn{3}{c|}{\textit{Regression}}\\
\cline{2-4}
MSE &  0.547 $\pm$ 0.003 & 0.596 $\pm$ 0.003 & N/A \\
Unfairness &   0.378 $\pm$ 0.007&\textbf{0.019 $\pm$ 0.005}  & N/A \\
Time (s) & N/A  &8.981 $\pm$ 1.319  & N/A \\
\hline
& \multicolumn{3}{c|}{\textit{Classification, Income - one sensitive}}\\
\cline{2-4}
 F1& 0.753$\pm$ 0.001 &0.737 $\pm$ 0.001  & 0.73  $\pm$ 0.002 \\
 Unfairness &0.170$\pm$ 0.001 &\textbf{0.003 $\pm$ 0.002}  & 0.021  $\pm$ 0.002 \\
  Time (s) & N/A &6.319 $\pm$ 0.422  & 100.8  $\pm$ 10.467 \\
 & \multicolumn{3}{c|}{\textit{Classification, Income - two sensitive}}\\
\cline{2-4}
 F1 &0.753$\pm$ 0.001&0.73 $\pm$ 0.001  & 0.73  $\pm$ 0.002 \\
 Unfairness &0.354$\pm$ 0.006& \textbf{0.009 $\pm$ 0.005}  & 0.207  $\pm$ 0.005 \\
 Time (s) & N/A &10.661 $\pm$ 0.140  & 355.275  $\pm$ 6.346 \\
 & \multicolumn{3}{c|}{\textit{Classification, Coverage}}\\
\cline{2-4}
 F1 &0.587 $\pm$ 0.0  &0.584 $\pm$ 0.0  & 0.574  $\pm$ 0.0 \\
 Unfairness &0.127 $\pm$ 0.0  &\textbf{0.011 $\pm$ 0.0}  & 0.119  $\pm$ 0.0 \\
 Time (s) & N/A &10.661 $\pm$ 0.147  & 355.2  $\pm$ 6.689 \\
 \hline
  & \multicolumn{3}{c|}{\textit{Benchmarks}}\\
& \textbf{\textit{FairBalance}} & \textbf{\textit{Our Method}} & \textbf{\textit{Fairlearn}} \\

 & \multicolumn{3}{c|}{\textit{\texttt{Bank} (n=45211,s=4)}}\\
\cline{2-4}
 F1 & 0.94$\pm$0.0 & 0.90$\pm$0.0 & 0.91$\pm$0.0 \\
 Unfairness & 0.16$\pm$0.03 &  \textbf{0.03$\pm$0.02} & \textbf{0.03$\pm$0.01} \\
 Time (s) & 2.7$\pm$1  & 1.2$\pm$0 & 83.1$\pm$21 \\
& \multicolumn{3}{c|}{\textit{\texttt{Adult} ($n=46447,s=2$)}}\\
\cline{2-4}
 F1 & 0.92 $\pm$0.0 & 0.94 $\pm$ 0.0 & 0.95 $\pm$ 0.0 \\
 Unfairness & 0.18$\pm$0.01 & \textbf{0.04$\pm$0.02} & 0.08 $\pm$0.02\\
 Time (s) & 2.6$\pm$1 & 1.1 $\pm$0.0 & 92.4 $\pm$ 33 \\
& \multicolumn{3}{c|}{\textit{\texttt{tunadromd} ($n=4464,s=10$)}}\\
\cline{2-4}
 F1 & 0.99$\pm$0.0 & 0.89$\pm$0.01 & 0.85$\pm$0.08 \\
 Unfairness & 3.77 $\pm$1.16 & \textbf{1.05$\pm$0.12} & 1.593$\pm$1.69 \\
 Time (s) & 0.8$\pm$0.5 & 0.2$\pm$0.1 & 24.5 $\pm$18 \\
\hline
\end{tabular}

\end{center}
\caption{Results for the correction of the biases for specific and benchmarking experiments.
}
\label{tab1}
\end{table}

Continuing with our application example (see Figure \ref{fig:application_all_figs}), our data reveals that the median income for Male participants is 17,000\$ higher than for Female participants. Similarly, the sensitive race's members have incomes 10,000\$ lower than the rest of the population. While not the sole contributor, these sub-groups could gain from fair predictions. 

For a decision maker, even when focusing solely on exact fairness, there are multiple routes to achieving fairness across both features. In our example, one could prioritize fairness in race scores first, followed by gender, or vice versa. Each step involves a performance loss but a fairness gain, depicted in the left panel of Figure \ref{fig:application_all_figs}. Although theory and the Figure demonstrate that the final outcome will be identical either way, practical implementation in steps presents a dilemma. The center panel of Figure \ref{fig:application_all_figs} displays our method's average corrective effects on predictions for two subgroups. For women in the sensitive racial group, correcting for race and gender boosts their predicted income. However, this poses a problem if both effects aren't corrected simultaneously. Women not in the sensitive racial group may oppose race-gender correction order due to a net deficit from race-only corrections. Conversely, sensitive subgroup women gain more from race corrections than gender, favoring the race-gender order. Our methodology offers flexibility for multiple fairness constraints. For instance, instead of consecutively rendering predictions fair on each feature, simultaneous steps are possible. In our example, if fair race predictions are sought without disadvantaging women, adjusting $\varepsilon$ for gender to one-sixth of race's size maintains average women's predictions while correcting for race. This adaptable approach is illustrated in the right panel of Figure \ref{fig:application_all_figs}, highlighting our methodology's strength. It ensures exact fairness results remain consistent regardless of the order in which scores become fair, while enabling decision makers to analyze stepwise fairness implementation effects.

\section{Conclusion}
We proposed a framework that expands the standard concept of fair scores from SSA to MSA. This extension ensures exact sequential fairness yields the same predictions regardless of correction order for the sensitive features. However, intermediate solutions can yield significant subgroup differences influenced by sensitive features. Our approach quantifies these differences and offers a way to mitigate them using approximate solutions. Our analysis raises intriguing research questions, including how optimal solutions change with fairness metrics beyond the DP constraint, particularly in the case of label-conditional metrics like equalized odds. 
The flexibility and user-friendliness of our method, supporting a comprehensive fairness approach across multiple debiasing steps, promotes the adoption of fair decision-making practices.


\section{Acknowledgments}
AC acknowledges Canada's National Sciences and Engineering Research Council (NSERC) for funding (RGPIN-2019-07077) and AC and PR  acknowledge funding from the SCOR Foundation for Science.
This research was enabled in part by support provided by Calcul Quebéc (calculquebec.ca) and the Digital Research Alliance of Canada (alliancecan.ca).

\bibliography{aaai24}


\appendix

\section{Supplementary Materials}

The supplementary material consists of two parts. One part contains all the proofs of our results, while the other part deals with additional numerical considerations. 

\subsection{A. Proof of main results}



Before providing the proof of the $\mathcal{W}_2$-barycentric associativity, we consider the following classical result in optimal transport theory in univariate measures \cite{santambrogio2015optimal, agueh2011barycenters}.

\begin{lem}\label{lem:OTuni}
    Let $\nu_1, \dots, \nu_K$ be $K$ univariate probability measures admitting densities, for all $w_1, \dots, w_K \geq 0$ summing to 1, the CDF of the optimal measure $\nu_{1:K} := {\rm Bar}(w_i, \nu_i)_{i=1}^K$ is given by
    $$
    F_{\nu_{1:K}}(\cdot) = \left( \sum_{i=1}^K w_i Q_{\nu_i} \right)^{-1} (\cdot)\enspace.
    $$
    Equivalently, the associated quantile function is given by
    $$
    Q_{\nu_{1:K}}(\cdot) = \left(\sum_{i=1}^K w_i Q_{\nu_i} \right)(\cdot)\enspace.
    $$ 
\end{lem}

\paragraph{Proof of Proposition \ref{prop:MainPropBary}} 
Given $K = K_1 + K_2 + \dots + K_r$ and the sets:
    $$
     B_1 = (w_{1, k}, \nu_{1, k})_{k=1}^{K_1}, \ldots, B_r = (w_{r, k}, \nu_{r, k})_{k=1}^{K_r}\enspace,$$
     where $\{w_{i, k}\}$ are positive and non-zero weights summing to 1 and $\{\nu_{i,k}\}$ represent univariate measures. In this context, the quantile function of the overall Wasserstein barycenter ${\rm Bar}\left\{B\right\}$, where $B :=B_1 \cup \cdots \cup B_r$ is given by (see above Lemma),
    \begin{align*}
        Q_{\nu_{B}}(u) &= \sum_{i=1}^{r}\sum_{k=1}^{K_i}w_{i,k}Q_{\nu_{i,k}} (u)\\
        &= \sum_{i=1}^{r}\underbrace{\left(\sum_{k'=1}^{K_i}w_{i,k'}\right)}_{=:\Tilde{w}_i} \underbrace{\left(\sum_{k=1}^{K_i}\frac{w_{i,k}}{\sum_{k'=1}^{K_i}w_{i,k'}}Q_{\nu_{i,k}}(u)\right)}_{=:Q_{\nu_{\Tilde{B}_i}}(u)} \\
        &= \sum_{i=1}^{r}\Tilde{w}_i Q_{\nu_{\Tilde{B}_i}}
    \end{align*}
where $Q_{\nu_{\Tilde{B}_i}}(u):=\sum_{k=1}^{K_i}\frac{w_{i,k}}{\Tilde{w}_i} Q_{\nu_{i,k}}(u)$ corresponds to the quantile function of the measure 
$$
\nu_{\Tilde{B}_i} := {\rm Bar}\left(\frac{w_{i,k}}{\Tilde{w}_i}, \nu_{i,k}\right)_{k=1, \ldots, K_i}\enspace.
$$
Thus,
\begin{equation*}
    {\rm Bar}\left(B_i \cup \cdots \cup B_r\right\} = {\rm Bar}\left(\Tilde{w}_{i}, \nu_{\Tilde{B}_i}\right)_{i=1, \dots, r}\enspace,
\end{equation*}
which concludes the proof.


Recall that $f^*(\boldsymbol{X}, \boldsymbol{A}) = \mathbb{E}[Y | \boldsymbol{X}, \boldsymbol{A}]$ and let us define the excess risk as $\mathcal{E}(\mathcal{H}) = \mathcal{R}(f^*) -  \inf_{h\in\mathcal{H}}\mathcal{R}(h)$ with $\mathcal{H}\subset \mathcal{F}$ a subclass of regressors. The following lemma is adapted from \cite{gouic2020projection} where we consider multiple sensitive attributes.

\begin{lem}\label{lem:ERisk}
Let $\mathcal{H}\subset\mathcal{F}$ be a subclass of regressors. If for any $h\in\mathcal{H}$ and $f\in\mathcal{F}$, where,
    $$
    \nu_{f|\boldsymbol{a}} = \nu_{h|\boldsymbol{a}} \quad \text{for all } \boldsymbol{a}\in\mathcal{A}\enspace,
    $$
    we have $f\in\mathcal{H}$, then the excess-risk of $\mathcal{H}$ is expressed as,
    \begin{equation*}
        \mathcal{E}(\mathcal{H}) = \inf_{h \in \mathcal{H}}\sum_{s\in\mathcal{S}}p_s \mathcal{W}_2^2\left(\nu_{f^*|\boldsymbol{a}}, \nu_{h|\boldsymbol{a}}\right)\enspace.
    \end{equation*}

\paragraph{Proof of Lemma \ref{lem:ERisk}}
From Pythagoras' Theorem, we can derive directly:
\begin{multline*}
     \mathbb{E}[(Y - h(\boldsymbol{X}, \boldsymbol{A}))^2] = \mathbb{E}[(Y-f^*(\boldsymbol{X}, \boldsymbol{A}))^2] \\ + 
     \mathbb{E}[(h(\boldsymbol{X}, \boldsymbol{A}) - f^*(\boldsymbol{X}, \boldsymbol{A}))^2]\enspace,
\end{multline*}
and therefore
\begin{multline*}
 \inf_{h \in \mathcal{H}} \mathbb{E}\big[\mathbb{E}[(h(\boldsymbol{X}, \boldsymbol{A}) - f^*(\boldsymbol{X}, \boldsymbol{A}))^2 | \boldsymbol{A}]\big] \\= \inf_{h \in \mathcal{H}} \mathbb{E}[(Y - h(\boldsymbol{X}, \boldsymbol{A}))^2] - \mathbb{E}[(Y - f^*(\boldsymbol{X}, \boldsymbol{A}))^2]\enspace.
\end{multline*}
From the definition of the Wasserstein distance, we have
\begin{align*}
    \mathbb{E}[(h(\boldsymbol{X}, \boldsymbol{A}) - f^*(\boldsymbol{X}, \boldsymbol{A}))^2|\boldsymbol{A}] \geq \mathbb{E}[\mathcal{W}_2^2(\nu_{f^*|\boldsymbol{A}}, \nu_{h | \boldsymbol{A}})]\enspace.
\end{align*}
Finally, since Assumption~\ref{assu:general} holds,
there also exists an optimal transport map $T_{\boldsymbol{a}}:\mathbb{R}\to \mathbb{R}$ \textit{s.t.} $T_{\boldsymbol{a}}\circ f^*(\boldsymbol{X}, \boldsymbol{a}) \sim \nu_{h | \boldsymbol{a}}$ and
\begin{align*}
     \mathcal{W}_2^2(\nu_{f^*|\boldsymbol{a}}, \nu_{h | \boldsymbol{a}}) &= \mathbb{E}[(T_{\boldsymbol{a}}\circ f^*(\boldsymbol{X}, \boldsymbol{a}) - f^*(\boldsymbol{X}, \boldsymbol{a}))^2]\enspace.
\end{align*}
Since $f\in\mathcal{H}$ for any $(\nu_{f|\boldsymbol{a}})_{\boldsymbol{a}} = (\nu_{h|\boldsymbol{a}})_{\boldsymbol{a}}$, it implies that 
\begin{align*}
    \mathbb{E}[\mathcal{W}_2^2(\nu_{f^*|\boldsymbol{A}}, \nu_{h | \boldsymbol{A}})] & = \mathbb{E}[(T_{\boldsymbol{A}}(f^*(\boldsymbol{X}, \boldsymbol{A})) - f^*(\boldsymbol{X}, \boldsymbol{A}))^2] \\ 
    & \geq \inf_{h \in \mathcal{H}} \mathbb{E}[(h(\boldsymbol{X}, \boldsymbol{A}) - f^*(\boldsymbol{X}, \boldsymbol{A}))^2]\enspace,
\end{align*}
which concludes the proof of the lemma.

\end{lem}

\paragraph{Proof of Proposition \ref{prop:MainFairOptimal}} Considering Lemma~\ref{lem:ERisk}, as we consider $\mathcal{F}_{\mathcal{A}-\text{fair}}$ of the form:
$$
\mathcal{F}_{\mathcal{A}-\text{fair}} := \left\{ f\in\mathcal{F} : \mathcal{U}(f) = 0 \right\}\enspace,
$$
representing the subclass of DP-fair predictors, we can easily infer that if
\begin{equation*}
    f_B = \argmin{f\in\mathcal{F}_{\mathcal{A}-\text{fair}}} \mathcal{R}(f)\enspace,
\end{equation*}
then it follows, directly from Lemma~\ref{lem:ERisk}, that $\nu_{f_B} = \mu_{\mathcal{A}}(\nu_{f^*})$. This, combined with Lemma~\ref{lem:OTuni}, furnishes us with the explicit closed-form solution:
\begin{equation*}
    f_B(\boldsymbol{x}, \boldsymbol{a}) = \left( \sum_{\boldsymbol{a}'\in\mathcal{A}} p_{\boldsymbol{a'}}Q_{f^*|\boldsymbol{a}'}\right)\circ F_{f^*|\boldsymbol{a}}\left( f^*(\boldsymbol{x}, \boldsymbol{a}) \right)\enspace,
\end{equation*}
concluding the proof of Proposition~\ref{prop:MainFairOptimal}.

Note that Proposition~\ref{prop:SFairOptimal} directly follows from Proposition~\ref{prop:SFairOptimalApprox}.

\paragraph{Proof of Propositions \ref{prop:SFairOptimal} and \ref{prop:SFairOptimalApprox}} 
\textit{w.l.o.g.}, we focus on the scenario that involves two sensitive characteristics, $\mathcal{A} = \mathcal{A}_1\times \mathcal{A}_2$. Leveraging Proposition~\ref{prop:MainPropBary} and the sets
\begin{align*}
    B_{1, 2} &= \left( \mathbb{P}(A_1 = a_1, A_2 = a_2), \nu_{f^*|a_1, a_2} \right)_{(a_1, a_2)\in\mathcal{A}_1\times \mathcal{A}_2}\\
    B_{1|a_2} &= \left( \mathbb{P}(A_1 = a_1| A_2 = a_2), \nu_{f^*|a_1, a_2}\right)_{a_1\in\mathcal{A}_1}\ \text{for } a_2\in\mathcal{A}_2\\
    B_{2|a_1} &= \left( \mathbb{P}(A_2 = a_2| A_1 = a_1), \nu_{f^*|a_2, a_1}\right)_{a_2\in\mathcal{A}_2}\ \text{for } a_1\in\mathcal{A}_1
\end{align*}
together with the Bayes theorem, we have, 
\begin{multline*}
    {\rm Bar}(B_{1,2}) = {\rm Bar}(\mathbb{P}(A_2 = a_2), {\rm Bar}(B_{1|a_2}))_{a_2\in\mathcal{A}_2}\\
    = {\rm Bar}(\mathbb{P}(A_1 = a_1), {\rm Bar}(B_{2|a_1}))_{a_1\in\mathcal{A}_1}
    \enspace.
\end{multline*}
Therefore,
$$
\mu_{\mathcal{A}}(\nu_{f^*}) = \mu_{\mathcal{A}_2}\circ \mu_{\mathcal{A}_1}(\nu_{f^*}) = \mu_{\mathcal{A}_1}\circ \mu_{\mathcal{A}_2}(\nu_{f^*})\enspace,
$$
which concludes the proof of Proposition~\ref{prop:SFairOptimal}.

Given the definition of the Unfairness measure in the MSA scenario as a sum of unfairness measures pertaining to individual sensitive attributes (refer to Eq.~\eqref{eq:UnfairnessMulti}), Proposition \ref{prop:SFairOptimalApprox} can be directly derived from Proposition 4.1 in \cite{chzhen2022minimax}, in conjunction with Proposition \ref{prop:SFairOptimal}.


\subsection{B. Additional numerical experiments}

\begin{figure}[tbhp]
\centerline{\includegraphics[scale=0.26]{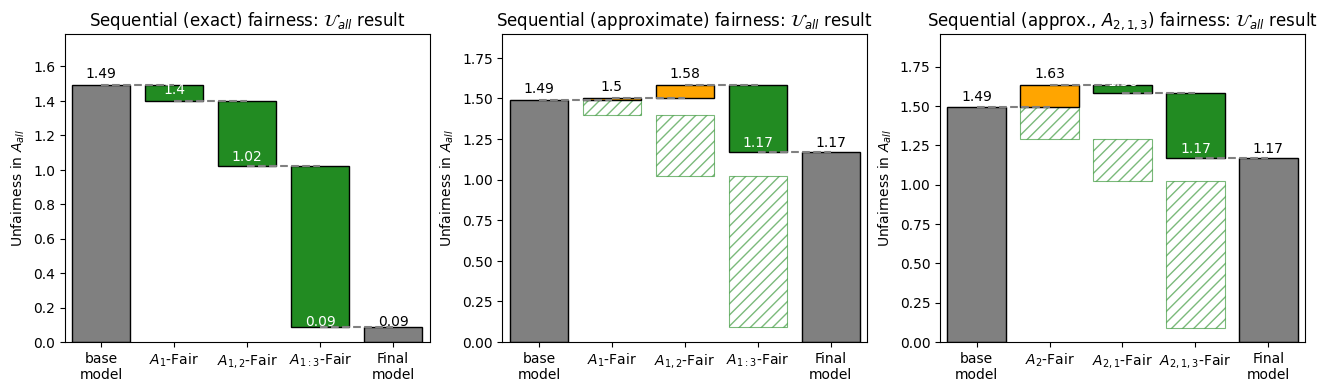}}
\caption{Synthetic data with parameter $\tau = (0, 0.05, 0.1)$. A sequential unfairness evaluation, $\mathcal{U} = \mathcal{U}_{1:3}$, of (\textit{left pane}) exact fairness, (\textit{middle pane}) approximate $A_{1, 2, 3}$-fairness with $\boldsymbol{\varepsilon}$-RI where $\boldsymbol{\varepsilon} = \varepsilon_{1, 2, 3} = (0.2, 0.5, 0.75)$ and (\textit{right pane}) approximate $A_{2, 1, 3}$-fairness with $\varepsilon_{2, 1, 3}$-RI. Hashed color corresponds to exact fairness.}
\label{fig:waterfall_syn_unfairnessAall}
\end{figure}

This section extends the analysis using synthetic data from the main body. Figure~\ref{fig:waterfall_syn_unfairnessAall} presents $\Hat{\mathcal{U}}_{1:r}$, an overall unfairness measure for all sensitive attributes. The proposed sequential fair approach (Prop. \ref{prop:SFairOptimal}) enhances interpretability by addressing inter-correlations among sensitive features while striving for fair predictions. Notably, this approach highlights the correlation of $A_1$ and $A_2$ with $A_3$. Pursuing (exact or approxiamte) fairness, as shown in Figure~\ref{fig:waterfall_syn_unfairnessAall}, indicates that making either $A_1$ or $A_2$ fair first impacts the overall unfairness differently in the initial step of the debiasing path. Specifically, making $A_2$ fair first reduces overall fairness, while making $A_1$ fair slightly enhances fairness. However, beyond this initial step, both paths eventually converge to the same performance.

\begin{figure}[tbhp]
\centerline{\includegraphics[scale=0.3]{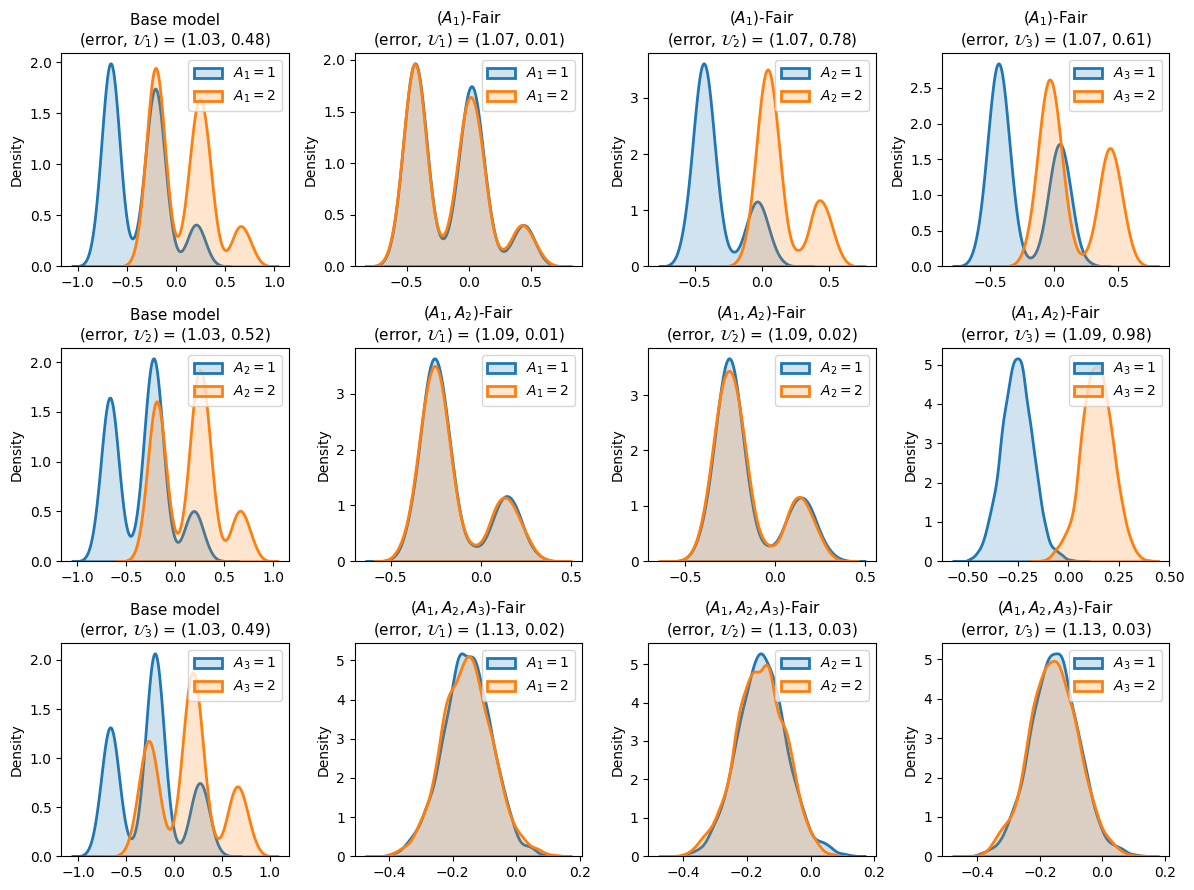}}
\caption{Synthetic data with parameters $\boldsymbol{\tau} = (0, 0.05, 0.1)$: sequential evaluation of exact fairness.}
\label{fig:kde_syn_unfairnessAall}
\end{figure}

\begin{figure}[tbhp]
\centerline{\includegraphics[scale=0.3]{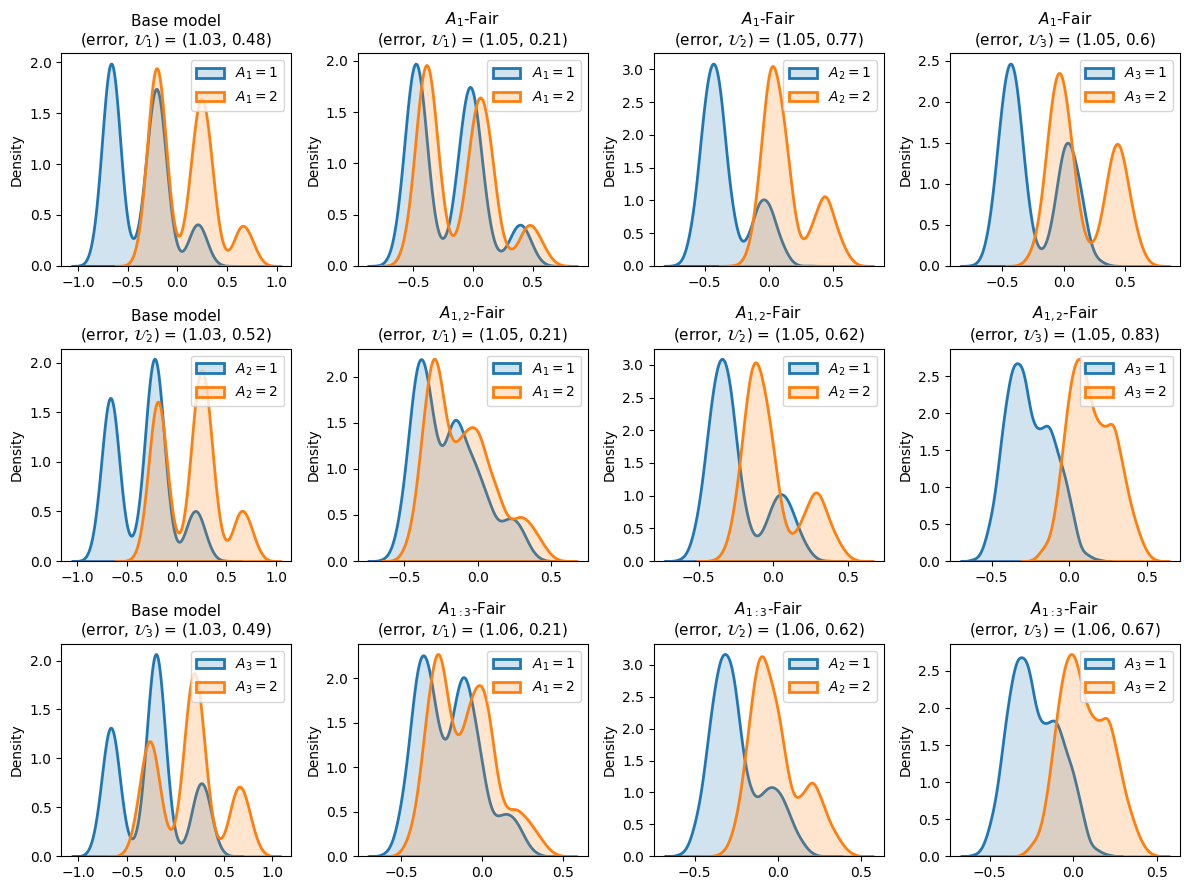}}
\caption{Synthetic data with parameters $\boldsymbol{\tau} = (0, 0.05, 0.1)$: evaluation of approximate fairness 
where $\boldsymbol{\varepsilon} = (0.2, 0.5, 0.75)$.}
\label{fig:kde_syn_unfairnessAall_approx}
\end{figure}

Figures~\ref{fig:kde_syn_unfairnessAall} and \ref{fig:kde_syn_unfairnessAall_approx} display intermediate solutions of our sequential fair mechanism on the synthetic data. Figure~\ref{fig:kde_syn_unfairnessAall} showcases exact fairness steps, while Figure~\ref{fig:kde_syn_unfairnessAall_approx} focuses on approximate fairness. Each row corresponds to enforcing fairness on an (additional) sensitive feature, while each column pertains to studying a specific sensitive feature. For example, the last (resp. first) row corresponds to enforcing fairness on $A_1$ (resp. $A_3$), and the last (resp. first) column corresponds to studying feature $A_1$ (resp. $A_3$). Both reveal intended score distribution differences, aligning perfectly (for exact fairness) or nearly so with some shifts (for approximate fairness), based on the chosen sensitive features (observed along the figures' diagonals).

\end{document}